\documentclass[11pt]{article}

\usepackage[T1]{fontenc}
\usepackage[utf8]{inputenc}
\usepackage[margin=1in]{geometry}

\usepackage{amsmath,amssymb,amsfonts}
\usepackage{bm}
\usepackage{mathtools}

\usepackage{graphicx}
\usepackage{booktabs}
\usepackage{multirow}
\usepackage{array}
\usepackage{tabularx}
\usepackage{ragged2e}
\usepackage{setspace}

\usepackage{algorithm}
\usepackage{algpseudocode}

\usepackage{xcolor}
\usepackage{microtype}

\usepackage{authblk}

\usepackage[hyphens]{url}
\usepackage[colorlinks=true,linkcolor=blue!50!black,citecolor=blue!50!black,urlcolor=blue!50!black]{hyperref}

\graphicspath{{./}}

\newcommand{\reftitle}[1]{\section*{#1}}
\newcommand{\acknowledgement}[1]{\vspace{12pt}\noindent\textbf{Acknowledgement:} #1\par}
\newcommand{\funding}[1]{\vspace{12pt}\noindent\textbf{Funding Statement:} #1\par}
\newcommand{\authorcontributions}[1]{\vspace{12pt}\noindent\textbf{Author Contributions:} #1\par}
\newcommand{\availabilityofdataandmaterials}[1]{\vspace{12pt}\noindent\textbf{Availability of Data and Materials:} #1\par}
\newcommand{\ethicsapproval}[1]{\vspace{12pt}\noindent\textbf{Ethics Approval:} #1\par}
\newcommand{\conflictsofinterest}[1]{\vspace{12pt}\noindent\textbf{Conflicts of Interest:} #1\par}

\newcommand{\keyword}[1]{\par\noindent\textbf{Keywords:} #1\par}

\title{BiLoG-Net: A Bi-Context Location-Guided Network for Breast Mass
Segmentation and Malignancy Classification in Mammography}

\author[1,5]{Abu Fatema Mohammad Abdun Noor}
\author[2,5]{Md Imam Ahasan}
\author[3]{Md Samiul Ahasan}
\author[4,*]{Kah Ong Michael Goh}
\author[1,*]{S M Hasan Mahmud}
\author[6]{Raihana Zannat}

\affil[1]{Department of Software Engineering, Daffodil International University, Dhaka, Bangladesh}
\affil[2]{College of Computer Science, Chongqing University, Chongqing, China}
\affil[3]{School of Computer Science and Technology, Xidian University, Xi'an, China}
\affil[4]{Center for Image and Vision Computing, COE for Artificial Intelligence, Faculty of Information Science \& Technology, Multimedia University, Jalan Ayer Keroh Lama, Melaka, 75450 Malaysia}
\affil[5]{BricksCloud AI, Uttara, Dhaka, Bangladesh}
\affil[6]{Department of Information and Communication Technology, Mawlana Bhashani Science \& Technology University, Bangladesh}
\affil[*]{Corresponding authors: Kah Ong Michael Goh (michael.goh@mmu.edu.my); S M Hasan Mahmud (drhasan.swe@diu.edu.bd)}

\date{}

\begin{document}
\maketitle

\begin{abstract}
Breast cancer remains the most commonly diagnosed malignancy among women worldwide, yet accurate detection and characterization of breast masses in mammography remain challenging due to subtle intensity variations, heterogeneous tissue densities, and indistinct lesion boundaries that complicate radiological interpretation. To address these limitations, we propose BiLoG-Net, a deep learning framework that jointly performs breast mass segmentation and malignancy classification through bi-context location-aware feature modeling and segmentation-guided attention mechanisms. Our architecture integrates a novel encoder-decoder paradigm with Fire-based feature extraction, lightweight global and local feature enhancement modules, and adaptive location-aware gating to simultaneously capture long-range contextual dependencies and fine-grained boundary-sensitive details. Unlike conventional multi-stage pipelines, our tightly coupled multi-task design enables mutual reinforcement between pixel-level localization and image-level diagnosis, reducing error propagation while producing spatially grounded malignancy predictions. Evaluated on CBIS-DDSM and INBreast benchmarks, BiLoG-Net achieves best performance among all CNN- and transformer-based baselines re-implemented and evaluated under our controlled experimental protocol, with Dice scores of 94.20\% and 93.10\%, classification accuracies of 95.20\% and 93.60\%, and AUC values of 97.10\% and 96.00\% on CBIS-DDSM and INBreast, respectively. By combining precise boundary delineation with reliable malignancy assessment in a single end-to-end model, this work holds strong potential for clinical computer-aided detection systems, helping radiologists prioritize suspicious cases and improve screening efficiency in busy clinical settings.
\end{abstract}

\keyword{Mammography; breast cancer detection; U-Net; multi-task learning; computer-aided diagnosis.}

\section{Introduction}
\label{introduction}

\begin{figure}[!ht]
    \centering
    \includegraphics[width=1\linewidth]{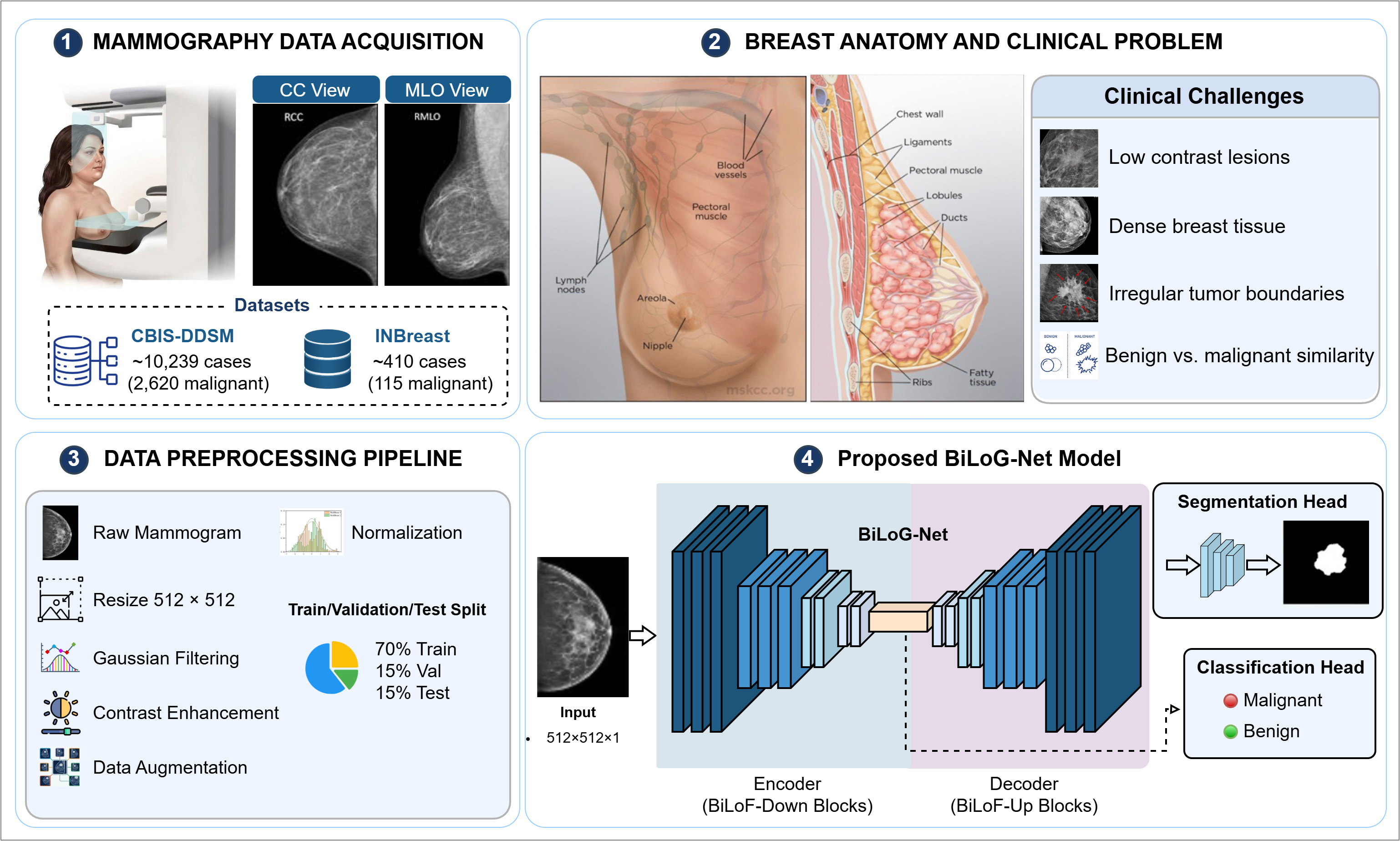}
    \caption{Overall workflow of the proposed BiLoG-Net framework for joint mammographic segmentation and classification. The figure illustrates the complete pipeline including mammography data acquisition, breast anatomy and clinical challenges, preprocessing steps, the proposed encoder-decoder architecture, and the final segmentation and malignancy classification outputs.}
    \label{fig:concept}
\end{figure}

The latest official data shows that breast cancer is now more common than lung cancer. Breast cancer is the primary cause of cancer mortality among females. By 2026 \cite{gurupakkiam2024survey}, breast, bowel, and lung cancers will be the most frequently diagnosed kinds of cancer in American women. Since peaking in 1989 \cite{sheshadri2005detection}, the incidence of breast cancer deaths in women has decreased by 43\%. The annual decrease in breast cancer mortality fell by approximately 2–3\% in the 1990 s to 1\% in 2024 \cite{ansari2024prediction}. Non-invasive breast cancer may remain localized inside a particular organ or structure, such as a duct or lobule, without spreading to distant tissues \cite{sohan2023systematic,hou2021anomaly}.
In contrast, malignant cells exhibit aberrant proliferation, infiltrating healthy tissues and inducing genetic mutations \cite{moo2018overview}. Tumors may be categorized into two distinct types: benign and malignant. Benign tumors are characterized by their inability to metastasize and are non-malignant. However, they might develop into masses due to excessive cell proliferation. On the other hand, malignant tumors are carcinogenic and can infiltrate nearby tissues and metastasize to different areas of the body. Breast cancer is the most commonly diagnosed cancer among women worldwide and a leading cause of cancer-related mortality. Globally, approximately 12.5\% of women are expected to be diagnosed with breast cancer during their lifetime. Breast cancer can originate in ducts, lobules, or surrounding tissues and may spread through glandular and lactiferous ductal structures, as illustrated in Figure~\ref{fig:concept}.

\begin{table*}[!t]
\centering
\footnotesize
\setlength{\tabcolsep}{3pt}
\renewcommand{\arraystretch}{1.15}

\caption{Comparison of state-of-the-art breast lesion segmentation and classification methods. Acc = Accuracy; Prec = Precision; Sens = Sensitivity; Spec = Specificity; Dice = Dice Similarity Coefficient; IoU = Intersection over Union; AUC = Area Under the Curve.}
\label{tab:comparison}

\resizebox{\textwidth}{!}{%
\begin{tabular}{p{1.0cm} p{4.5cm} p{3.0cm} p{8.0cm}}
\toprule
\textbf{Ref.} &
\textbf{Method} &
\textbf{Dataset} &
\textbf{Performance Metrics} \\
\midrule

\cite{ahmad2025multi}
& MSDLM (GaU-Net + EfficientNetV2-B0 + CNN)
& CBIS-DDSM, WBCD
& Acc = 97.6\%, IoU = 85.59\%, Prec = 94.5\%, Recall = 97.25\%, AUC = 95.75\% \\

\cite{mohammadi2025enhanced}
& HTU-Net (CNN + Transformer U-Net)
& CBIS-DDSM, INbreast
& Dice = 93.50\% (CBIS-DDSM), 92.14\% (INbreast); Acc = 98.43\% \\

\cite{wang2025robust}
& Bi-CBMSegNet (ResNet-152 Backbone)
& DDSM, INbreast
& Dice = 97.09\% (DDSM), 97.81\% (INbreast); Acc = 99.39\%; AUC = 98.00\% \\

\cite{bilal2025quantum}
& Q-BGWO-SQSVM
& MIAS, INbreast, DDSM, CBIS-DDSM
& Acc = 98.0\%; Sens = 96.0\%; Spec = 99.0\% \\

\midrule

\cite{han2024deep}
& DLSEN-RS (DenseNet-169)
& CBIS-DDSM, INbreast
& AUC = 76.3\%/95.1\%; Acc = 74.2\%/83.7\% \\

\midrule

\cite{zhong2023a}
& Mix Transformer + U-Net
& INbreast, CBIS-DDSM
& Dice = 83.63\% (INbreast), 89.90\% (CBIS-DDSM) \\

\cite{iqbal2023a}
& Swin Transformer
& CBIS-DDSM
& Dice = 77.11\% \\

\cite{liu2023a}
& Transformer Encoder--Decoder
& CBIS-DDSM, INbreast
& Dice = 92.20\% (CBIS-DDSM), 91.83\% (INbreast) \\

\midrule

\cite{su2022a}
& YOLO + MedT
& CBIS-DDSM, INbreast
& F1-score = 74.5\%; IoU = 64.0\% \\

\midrule

\cite{salama2021a}
& Modified U-Net
& MIAS, CBIS-DDSM
& Acc = 98.87\%; AUC = 98.88\%; Sens = 98.98\%; Prec = 98.79\%; F1-score = 97.99\% \\

\cite{yu2021a}
& CrossoverNet
& DDSM, INbreast
& Dice = 92.50\% (DDSM), 91.26\% (INbreast) \\

\cite{rajalakshmi2021a}
& DS U-Net + Dense CRFs
& DDSM, INbreast
& Dice = 82.9\% (DDSM), 79.0\% (INbreast) \\

\midrule

\cite{yu2020a}
& Dense Mask R-CNN
& CBIS-DDSM
& Prec = 65.0\% \\

\cite{min2020a}
& Mask R-CNN
& INbreast
& Dice = 88.0\% \\

\cite{ahmed2020a}
& DeepLab + Mask R-CNN
& MIAS, CBIS-DDSM
& AUC = 98.0\%/95.0\%; Prec = 80.0\%/75.0\% \\

\cite{bhatti2020a}
& Mask R-CNN-FPN
& DDSM, INbreast
& Prec = 84.0\%; Acc = 91.0\% \\

\cite{chen2020a}
& Modified U-Net
& INbreast, CBIS-DDSM
& Dice = 81.64\%/82.16\%; Acc = 99.43\%/99.81\% \\

\cite{sun2020a}
& AUNet
& CBIS-DDSM, INbreast
& Dice = 81.8\% (CBIS-DDSM), 79.1\% (INbreast) \\

\cite{abdelhafiz2020a}
& Vanilla U-Net
& Private Dataset
& Acc = 92.6\%; Dice = 95.1\%; IoU = 90.9\% \\

\midrule

\cite{wang2019a}
& PyramidNet
& INbreast, DDSM
& Dice = 91.10\% (DDSM), 91.69\% (INbreast) \\

\cite{li2019a}
& Dense U-Net + Attention Gates
& DDSM
& F1-score = $82.24\pm0.06$; Sens = $77.89\pm0.08$ \\

\cite{shen2019b}
& MS-ResCU-Net
& INbreast
& Acc = 94.16\%; Dice = 91.78\% \\

\midrule

\cite{li2018a}
& CRU-Net
& INbreast, DDSM-BCRP
& Dice = 93.66\% (INbreast), 93.32\% (DDSM-BCRP) \\

\bottomrule
\end{tabular}}
\end{table*}

Mammography has been crucial in the timely identification of breast cancer, leading to a considerable decrease in death rates. This is achieved by capturing breast images from two angles: Medio lateral oblique and craniocaudally. This technique classifies breast density into four categories: adipose, dispersed, heterogeneously dense, and very dense \cite{momenimovahed2019epidemiological,lopez2017breast}. Tumors exhibit a range of forms and borders, with malignant tumors often displaying irregular and unclear boundaries, whereas benign tumors are frequently compact and well-defined. Calcifications exhibit various characteristics; non-cancerous calcifications usually manifest as sizable rods or popcorn-shaped structures, while cancerous ones are spread out or grouped \cite{kretz2020mammography}. Multiple factors contribute to the formation of breast cancer. 
Another key challenge in mammogram classification is that mammographic images are vastly different from natural images, which implies that the mere application of a proficient deep learning model designed for natural images to mammograms would yield suboptimal results, falling short of expectations and requirements within the context of diagnostic imaging. In terms of density, breast mass is usually of equal density and therefore has characteristic pixel intensities that are not easily distinguishable. In shape, breast mass can be oval, irregular, or round with multiple border shapes such as needle-like, square, blurry, or unclear \cite{oliver2010review}. However, the diagnosis of breast lumps is strongly linked to their shape, contrast, and boundaries, which makes lump classification a difficult challenge. At the same time, due to their complex and diverse appearance, breast lesions are difficult to detect and classify even by professional radiologists. Furthermore, the diagnosis of the whole mammographic image is determined by the condition of the lesion region, which is only an extremely small part of a whole image \cite{dhungel2017fully, zhu2017deep, lotter2017multi}. CNN methods currently used in the field of computer vision usually use the standard inputs of 299×299 or 224×224 \cite{carneiro2015unregistered, zhu2017deep}. This means that for a typical CNN, maintaining the mammographic image size requires more computing power and training time, while shrinking the mammographic image will result in information loss and reduced prediction accuracy.

To address the limitations of existing mammographic analysis frameworks, this work proposes a novel multi-task deep learning architecture termed BiLoG-Net for joint breast mass segmentation and malignancy classification. Figure~\ref{fig:concept} presents the overall workflow of the proposed framework, including mammography data acquisition, preprocessing, bi-context feature modeling, segmentation, and malignancy classification. The primary contributions of this work are summarized as follows:

\begin{itemize}
    \item We propose a novel Bi-Context Location-Aware Feature Modeling framework 
          that integrates depth-wise adaptive bi-context fusion (learning optimal 
          global-local balance per encoder stage) with a residual-preserving 
          location-aware gate designed to avoid suppressing subtle low-contrast 
          lesion signals.
    
    \item We introduce a unified encoder-decoder architecture with tight end-to-end 
          coupling of segmentation-guided classification under a single joint-loss 
          framework, enabling mutual reinforcement between pixel-level localization 
          and image-level diagnosis.
    
    \item Extensive experiments conducted on the CBIS-DDSM and INBreast benchmark 
          datasets demonstrate that BiLoG-Net consistently outperforms existing 
          CNN- and transformer-based baselines across multiple segmentation and 
          classification metrics.
\end{itemize}

\section{Related Work}
Recent advances in mammographic image analysis have explored both multi-stage pipelines and hybrid architectures to improve breast lesion segmentation and classification. Ahmad et al.\ proposed a multi-stage deep learning model (MSDLM) that first employs a Gaussian U-Net (GaU-Net) to segment suspicious regions, then uses EfficientNetV2-B0 to extract high-level features from the segmented lesions, and finally applies a CNN classifier to distinguish benign from malignant cases, achieving an accuracy of about 97.6\% with high IoU and AUC on CBIS-DDSM and tabular benchmarks \cite{ahmad2025multi}. To further enhance mass delineation, a hybrid transformer U-Net (HTU-Net) integrates CNN-based encoder-decoder features with multiscale cross-attention transformers, capturing both local details and long-range dependencies; evaluated on CBIS-DDSM and INbreast, this model reports Dice scores above 92\% and consistently outperforms classical U-Net and recent transformer-based baselines \cite{mohammadi2025enhanced}. Complementing these efforts, the Bi-CBMSegNet framework introduces a dual-module encoder–decoder with Global Feature Enhancement (GFEM) and Local Feature Enhancement (LFEM) modules, enabling robust modeling of homogeneous and heterogeneous breast tissue; on DDSM and INbreast, Bi-CBMSegNet attains Dice coefficients around 97\% with accuracy exceeding 99\%, surpassing several state-of-the-art segmentation networks \cite{wang2025robust}.

Beyond segmentation-focused approaches, recent work has also targeted end-to-end classification directly from full mammograms. The Deep Location Soft-Embedding Network with Regional Scoring (DLSEN-RS) leverages a DenseNet backbone with a location soft-embedding module and region-based, group-max pooling, allowing the network to automatically discover discriminative regions using only image-level labels; this method achieves superior AUC and accuracy on INbreast and CBIS-DDSM compared with earlier ROI-based and multi-stage classifiers \cite{han2024deep}. In parallel, a quantum-optimized SqueezeNet–SVM framework (Q-BGWO-SQSVM) couples deep features extracted by SqueezeNet with an SVM whose hyperparameters are tuned via a Quantum Binary Grey Wolf Optimizer, yielding highly competitive performance across MIAS, INbreast, DDSM, and CBIS-DDSM, with reported accuracies up to 98\%, sensitivities around 96\%, and specificities near 99\% under cross-validation \cite{bilal2025quantum}. Together, these studies demonstrate that carefully designed segmentation backbones, attention mechanisms, and advanced optimization or pooling strategies can substantially boost the reliability of automated breast cancer detection in mammography.

A study in Ref.~\cite{bhatti2020multi} introduced a Masked Regional Convolutional Neural Network with a Feature Pyramid Network (Mask R-CNN-FPN), further enhancing feature representation and accuracy. The U-Net architecture and its many derivative models continue to be widely used and favored approaches for breast mass segmentation. Studies~\cite{li2018improved, chen2020novel, salama2021deep} proposed various modifications to the U-Net architecture for better detail capture and improved segmentation accuracy. The robustness of U-Net across multiple medical tasks was demonstrated in Ref.~\cite{abdelhafiz2018convolutional}, and a model called CrossoverNet was introduced in Ref.~\cite{yu2021crossover} to enhance feature extraction through crossover learning. Yu et al. introduced CrossoverNet to enhance feature extraction through crossover learning. U-Net models are advantageous for their capability to preserve spatial information through skip connections, making them particularly effective for medical segmentation. However, standard U-Net models can struggle with capturing finer details on complex datasets, often requiring enhancements to maintain high resolution in segmented areas. Pyramid networks like PyramidNet~\cite{wang2019multi}, expand U-Net’s ability to capture diverse feature scales for improved accuracy, while attention mechanisms proposed in Refs.~\cite{li2019attention, sun2020aunet} further refine segmentation by focusing on relevant areas, though they may introduce additional complexity and computational costs.

Hybrid models that combine multiple methods offer unique strengths. Studies~\cite{ramesh2021deeply} introduced methods like MS-ResCU-Net and DS U-Net, use multiple levels of supervision and dense CRFs to improve segmentation precision. Hybrid models benefit from integrating the strengths of different techniques, such as context-awareness from transformers and localization from CNNs, making them robust and versatile. However, these models can be computationally intensive and challenging to balance, requiring careful tuning to avoid overfitting. In ~\cite{xun2022rga}, the authors proposed a hybrid model that combines both ResNet and InceptionV2 for detecting breast tumors from the scanned input images. The CVC Clinic-DB dataset was used to test the performance of this hybrid model, achieving an accuracy of 91.2\%, which is a good output compared to other existing proposed models. If we see all these existing models, the performance of the individual model is much less compared to the performance of a hybrid model in the detection of tumors. Therefore, a hybrid model is necessary to achieve better accuracy in breast lesion identification and to obtain a good output within a faster execution time interval. The methods summarized in Table 1 are collected from the literature for contextual positioning only. Because these studies differ in dataset version, preprocessing pipeline, train–test split, and evaluation protocol, their reported values are not directly comparable to the controlled comparisons in Section 4, where all baselines are re-implemented and trained by us under an identical protocol. Beyond purely deep-learning pipelines, hybrid approaches that combine learned 
features with case-based reasoning (CBR) and other non-end-to-end paradigms have 
also been explored for mammogram classification. For example, a CNN-CBR system 
was proposed by Bouzar-Benlabiod and Harrar~\cite{bouzar2023novel}, which 
couples convolutional feature extraction with a case-retrieval classification 
mechanism, achieving competitive image-level classification performance. Unlike 
BiLoG-Net, this line of work targets image-level classification via 
similarity-based case retrieval and does not address pixel-level lesion 
segmentation, nor is it evaluated on the CBIS-DDSM or INBreast benchmarks used in 
this study; a direct, controlled comparison would therefore require 
reimplementing the CBR pipeline under our segmentation-and-classification 
protocol and datasets, which we identify as a valuable direction for future 
benchmarking alongside other non-deep-learning paradigms.

\begin{figure}[!ht]
    \centering
    \includegraphics[width=1\linewidth]{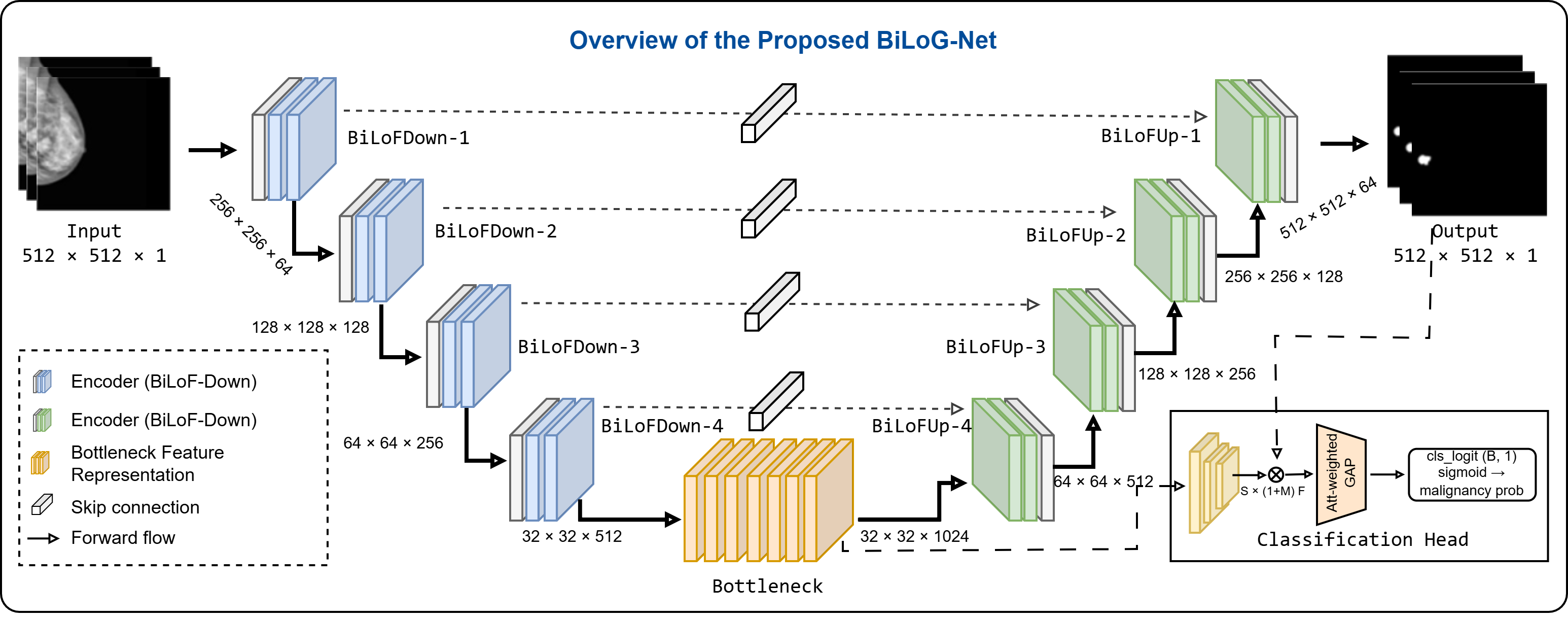}
    \caption{Overview of the proposed BiLoG-Net architecture for joint mammographic segmentation and malignancy classification. The framework employs a shared encoder-decoder structure with bi-context feature modeling, location-aware gating, gated skip connections, and segmentation-guided classification for accurate lesion localization and diagnosis.}
    \label{fig:proposed_model}
\end{figure}

\section{Method}
\subsection{Overview of the Framework}
\label{sec:overview_framework}

This work proposes  Bi-Context Location-Aware U-Net (BiLoG-Net) a deep learning framework, designed to jointly perform breast mass segmentation and malignancy classification from mammographic images. The proposed architecture follows an encoder-decoder paradigm inspired by U-Net, while introducing novel bi-context modeling and location-aware feature gating mechanisms to address the challenges of low contrast, heterogeneous tissue distribution, and ambiguous lesion boundaries commonly observed in mammograms.
Let $\mathbf{X} \in \mathbb{R}^{H \times W \times 1}$ denote an input grayscale mammogram image, where $H$ and $W$ represent the spatial dimensions. The network consists of a shared encoder $\mathcal{E}(\cdot)$, a bottleneck representation $\mathbf{F}$, and two task-specific heads: a segmentation decoder $\mathcal{D}(\cdot)$ and a classification head $\mathcal{C}(\cdot)$. The overall forward process of the proposed framework can be expressed as
\begin{equation}
\mathbf{F}, \{\mathbf{S}_l\}_{l=1}^{L} = \mathcal{E}(\mathbf{X}),
\end{equation} 
where $\mathbf{F} \in \mathbb{R}^{h \times w \times C}$ denotes the deep bottleneck feature map, and $\{\mathbf{S}_l\}_{l=1}^{L}$ represent multi-scale skip features extracted at different encoder depths. These skip features preserve fine-grained spatial information that is essential for accurate lesion boundary delineation.
The segmentation branch reconstructs a pixel-wise prediction map by progressively upsampling the bottleneck features while selectively incorporating encoder skip features through gated skip connections. Formally, the segmentation output is given by
\begin{equation}
\hat{\mathbf{Y}}_{\text{seg}} = \sigma \left( \mathcal{D}\left(\mathbf{F}, \{\mathbf{S}_l\}_{l=1}^{L}\right) \right),
\end{equation}
where $\hat{\mathbf{Y}}_{\text{seg}} \in [0,1]^{H \times W}$ is the predicted probability map for breast mass regions and $\sigma(\cdot)$ denotes the sigmoid activation function. The decoder employs bi-context refinement blocks that integrate global contextual cues and local boundary-sensitive features, enabling robust segmentation under varying tissue densities.
In parallel, the classification branch operates directly on the shared bottleneck representation $\mathbf{F}$, which encodes high-level semantic information relevant to malignancy assessment. To enhance diagnostic relevance, the framework introduces a segmentation-guided attention mechanism that leverages the predicted segmentation mask to emphasize lesion-centric regions during feature aggregation. Specifically, a spatial attention map $\mathbf{A}$ is generated as
\begin{equation}
\mathbf{A} = \sigma\left(\mathcal{W}_a * \mathbf{F}\right),
\end{equation}
where $\mathcal{W}_a$ denotes a $1\times1$ convolution kernel and $*$ represents the convolution operation. The attention-weighted global feature vector $\mathbf{v} \in \mathbb{R}^{C}$ is then computed as
\begin{equation}
\mathbf{v} = \frac{\sum_{i,j} \mathbf{F}_{i,j} \cdot \mathbf{A}_{i,j}}{\sum_{i,j} \mathbf{A}_{i,j} + \epsilon},
\end{equation}
with $(i,j)$ indexing spatial locations and $\epsilon$ ensuring numerical stability. The final classification output is obtained through a fully connected mapping
\begin{equation}
\hat{y}_{\text{cls}} = \mathcal{C}(\mathbf{v}),
\end{equation}
where $\hat{y}_{\text{cls}} \in [0,1]$ denotes the predicted probability of malignancy.
The proposed joint learning strategy enables effective information sharing between segmentation and classification tasks. The segmentation branch benefits from discriminative features learned for malignancy prediction, while the classification branch is guided by spatially precise lesion localization. This mutual reinforcement mitigates the limitations of single-task models and reduces error propagation commonly observed in multi-stage pipelines.
Figure~\ref{fig:proposed_model} illustrates the overall architecture of the proposed BiLoG-Net, highlighting the shared encoder, bi-context bottleneck, gated decoder for segmentation, and the segmentation-guided classification head. The unified design allows end-to-end optimization under a multi-task learning objective, resulting in improved robustness and generalization across diverse mammographic datasets.

\begin{figure}[!ht]
    \centering
    \includegraphics[width=0.8\textwidth]{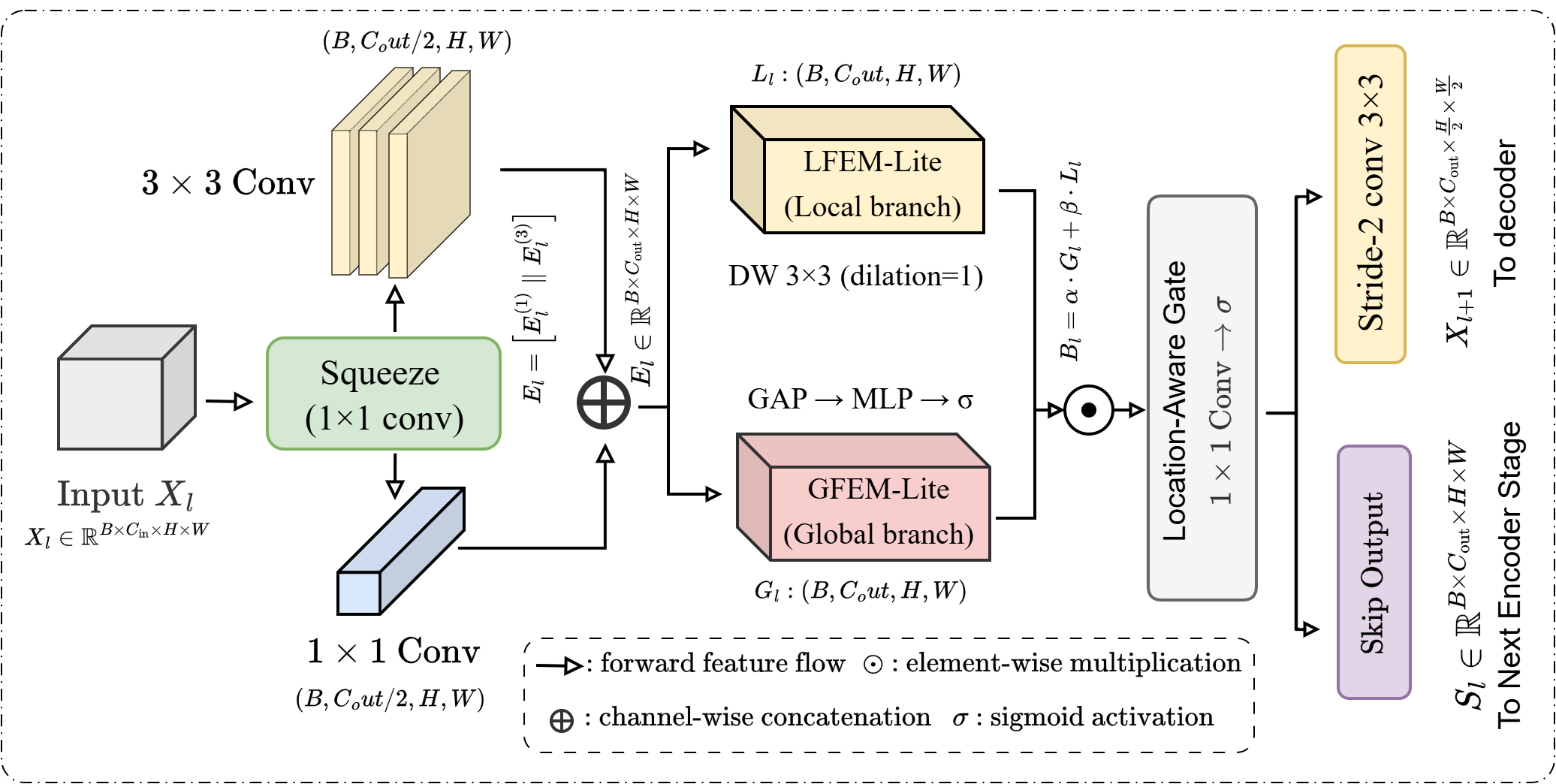}
    \caption{Architecture of the proposed BiLoF-Down block combining Fire-based feature extraction, global and local feature enhancement, adaptive bi-context fusion, and location-aware gating for mammographic analysis.}
    \label{fig:bilof_down_block}
\end{figure}

\subsection{Bi-Context Location-Aware Encoder (Downsampling Blocks)}
\label{sec:bilof_encoder}

The encoder of the proposed BiLoG-Net is designed to hierarchically extract discriminative representations from mammographic images while preserving both global contextual awareness and fine-grained local structural details. Each encoder stage is implemented using a novel \textit{Bi-Context Location-Aware Downsampling} (BiLoF-Down) block, which integrates Fire-based feature extraction, global and local feature enhancement modules, spatially adaptive location gating, and an efficient downsampling strategy. This design aims to address the intrinsic challenges of mammographic analysis, including subtle intensity variations, dense tissue overlap, and indistinct lesion boundaries.
Let $\mathbf{X}_l \in \mathbb{R}^{H_l \times W_l \times C_l}$ denote the input feature map to the $l$-th encoder stage. The BiLoF-Down block transforms $\mathbf{X}_l$ into a refined feature representation $\mathbf{X}_l^{\prime}$ and a downsampled output $\mathbf{X}_{l+1}$, which is propagated to the subsequent encoder stage.

\textbf{Fire-Based Feature Extraction:} To achieve computational efficiency without sacrificing representational capacity, the encoder employs a Fire-based feature extraction module inspired by SqueezeNet. The Fire block decomposes standard convolutional operations into a squeeze layer followed by parallel expand layers, enabling channel-wise compression and diversified spatial feature learning.
Formally, given an input feature map $\mathbf{X}_l$, the squeeze operation is defined as
\begin{equation}
\mathbf{Z}_l = f_{1\times1}(\mathbf{X}_l),
\end{equation}
where $f_{1\times1}(\cdot)$ denotes a $1\times1$ convolution that reduces the channel dimensionality. The squeezed features are then expanded using parallel $1\times1$ and $3\times3$ convolutions:
\begin{equation}
\mathbf{E}_l = \left[ f_{1\times1}(\mathbf{Z}_l) \; \Vert \; f_{3\times3}(\mathbf{Z}_l) \right],
\end{equation}
where $\Vert$ denotes channel-wise concatenation. This design enables the encoder to capture both cross-channel interactions and local spatial patterns critical for lesion characterization.

\textbf{Global Feature Enhancement (GFEM-Lite):} While local convolutions are effective for capturing fine details, mammographic interpretation often requires global contextual understanding to differentiate lesions from complex background tissue. To this end, a lightweight Global Feature Enhancement Module (GFEM-Lite) is introduced to adaptively recalibrate channel-wise feature responses.
Given the extracted feature map $\mathbf{E}_l$, global context is encoded via global average pooling:
\begin{equation}
\mathbf{g}_l = \frac{1}{H_l W_l} \sum_{i=1}^{H_l} \sum_{j=1}^{W_l} \mathbf{E}_l(i,j),
\end{equation}
where $\mathbf{g}_l \in \mathbb{R}^{C_l}$. The pooled descriptor is passed through a bottleneck transformation followed by a sigmoid activation to produce channel attention weights:
\begin{equation}
\mathbf{a}_l = \sigma \left( \mathbf{W}_2 \, \delta \left( \mathbf{W}_1 \mathbf{g}_l \right) \right),
\end{equation}
where $\mathbf{W}_1$ and $\mathbf{W}_2$ are learnable parameters, $\delta(\cdot)$ denotes the ReLU activation, and $\sigma(\cdot)$ is the sigmoid function. The globally enhanced feature map is then obtained as
\begin{equation}
\mathbf{G}_l = \mathbf{E}_l \odot \mathbf{a}_l,
\end{equation}
with $\odot$ representing channel-wise multiplication.

\textbf{Local Feature Enhancement (LFEM-Lite): } To complement global context modeling, the encoder incorporates a Local Feature Enhancement Module (LFEM-Lite) that focuses on boundary-sensitive and texture-specific information. LFEM-Lite utilizes depthwise separable convolutions to efficiently capture local spatial variations while maintaining low computational complexity.
The local enhancement operation is defined as
\begin{equation}
\mathbf{L}_l = f_{1\times1}\left( f_{3\times3}^{\text{dw}}(\mathbf{E}_l) \right),
\end{equation}
where $f_{3\times3}^{\text{dw}}(\cdot)$ denotes a depthwise $3\times3$ convolution and $f_{1\times1}(\cdot)$ is a pointwise convolution for channel mixing. This formulation enhances edge continuity and boundary localization, which are crucial for accurate mass segmentation.

\textbf{Bi-Context Feature Fusion:} The global and local enhanced features are adaptively fused to form a bi-context representation:
\begin{equation}
\mathbf{B}_l = \alpha \mathbf{G}_l + \beta \mathbf{L}_l,
\end{equation}
where $\alpha$ and $\beta$ are learnable scalar parameters that control the relative contribution of global and local context. This adaptive fusion allows the network to dynamically balance contextual awareness and spatial precision.

\textbf{Location-Aware Gating:} To further emphasize anatomically and diagnostically relevant regions, a location-aware gating mechanism is applied to the fused features. A spatial attention map is generated as
\begin{equation}
\mathbf{M}_l = \sigma \left( f_{1\times1}(\mathbf{B}_l) \right),
\end{equation}
where $\mathbf{M}_l \in [0,1]^{H_l \times W_l}$ represents the learned importance of each spatial location. The gated feature map is then computed as
\begin{equation}
\mathbf{X}_l^{\prime} = \mathbf{B}_l \odot (1 + \mathbf{M}_l),
\end{equation}
which reinforces lesion-prone regions while preserving background contextual information.

\textbf{Downsampling Strategy: }Finally, spatial resolution is reduced to enable hierarchical feature abstraction and increase the receptive field. The downsampling operation is implemented using a stride-2 convolution:
\begin{equation}
\mathbf{X}_{l+1} = f_{3\times3}^{s=2}(\mathbf{X}_l^{\prime}),
\end{equation}
which jointly performs feature transformation and resolution reduction. The pre-downsampled gated features $\mathbf{X}_l^{\prime}$ are preserved as skip connections for the decoder to facilitate precise spatial reconstruction.
Figure~\ref{fig:bilof_down_block} illustrates the internal structure of the BiLoF-Down block, highlighting the integration of Fire-based feature extraction, bi-context enhancement, location-aware gating, and downsampling. This encoder design enables the proposed framework to learn robust, context-aware representations that are well-suited for joint breast mass segmentation and classification.

\subsection{Bottleneck, Decoder, and Task-Specific Heads}
\label{sec:bottleneck_decoder_heads}

\textbf{Bottleneck Representation: }At the deepest level of the proposed architecture, a shared bottleneck representation is employed to aggregate high-level semantic information for both segmentation and classification tasks. This bottleneck serves as a critical junction that bridges the encoder and decoder while simultaneously acting as the primary feature source for malignancy prediction.
Let $\mathbf{X}_L \in \mathbb{R}^{h \times w \times C_L}$ denote the output feature map of the final encoder stage. The bottleneck transformation is defined as
\begin{equation}
\mathbf{F} = \phi_b(\mathbf{X}_L),
\end{equation}
where $\phi_b(\cdot)$ represents a sequence of convolutional and bi-context refinement operations. Specifically, the bottleneck incorporates global-local context aggregation to enhance semantic consistency and robustness across varying lesion appearances. By capturing both long-range dependencies and localized discriminative patterns, the bottleneck feature map $\mathbf{F}$ provides a compact yet expressive representation that is jointly optimized for pixel-level localization and image-level diagnosis.
The shared nature of the bottleneck facilitates effective multi-task learning, enabling gradient signals from both segmentation and classification objectives to influence the learned representation. This design encourages the extraction of features that are simultaneously spatially informative and diagnostically relevant.

\textbf{Decoder with Skip-Aware Refinement (BiLoF-Up Blocks): } The decoder reconstructs high-resolution segmentation maps by progressively upsampling the bottleneck features while integrating encoder skip connections through a skip-aware refinement mechanism. Each decoder stage is implemented using a BiLoF-Up block, which consists of an upsampling operation, gated skip-connection fusion, and contextual feature refinement.
Given a decoder input feature map $\mathbf{D}_{l+1} \in \mathbb{R}^{h_{l+1} \times w_{l+1} \times C_{l+1}}$, spatial resolution is first increased using bilinear interpolation:
\begin{equation}
\tilde{\mathbf{D}}_l = \mathcal{U}(\mathbf{D}_{l+1}),
\end{equation}
where $\mathcal{U}(\cdot)$ denotes a $2\times$ upsampling operation. A $1\times1$ convolution is subsequently applied to adjust the channel dimensionality.
To suppress irrelevant background information introduced by direct skip fusion, a skip-connection gating mechanism is employed. Given the corresponding encoder skip feature $\mathbf{S}_l$, a spatial gating map is generated from the upsampled decoder features:
\begin{equation}
\mathbf{G}_l = \sigma \left( f_{1\times1}(\tilde{\mathbf{D}}_l) \right),
\end{equation}
and applied to the skip feature as
\begin{equation}
\hat{\mathbf{S}}_l = \mathbf{S}_l \odot \mathbf{G}_l.
\end{equation}
This gating operation selectively preserves lesion-relevant spatial regions while reducing noise from normal tissue structures.
The gated skip feature $\hat{\mathbf{S}}_l$ is concatenated with the upsampled decoder feature and refined using bi-context enhancement:
\begin{equation}
\mathbf{D}_l = \phi_u \left( \left[ \tilde{\mathbf{D}}_l \; \Vert \; \hat{\mathbf{S}}_l \right] \right),
\end{equation}
where $\phi_u(\cdot)$ denotes a combination of global and local feature enhancement followed by Fire-based refinement. This process is repeated across decoder stages until the original input resolution is restored. 
\textbf{Segmentation Head: }The segmentation head operates on the final decoder output $\mathbf{D}_0 \in \mathbb{R}^{H \times W \times C_0}$ to generate a dense pixel-wise prediction map. A $1\times1$ convolution is employed to project the multi-channel feature map into a single-channel logit map:
\begin{equation}
\mathbf{Z}_{\text{seg}} = f_{1\times1}(\mathbf{D}_0).
\end{equation}
The final segmentation probability map is obtained using a sigmoid activation:
\begin{equation}
\hat{\mathbf{Y}}_{\text{seg}} = \sigma(\mathbf{Z}_{\text{seg}}),
\end{equation}
where $\hat{\mathbf{Y}}_{\text{seg}} \in [0,1]^{H \times W}$ represents the predicted likelihood of each pixel belonging to a breast mass. This formulation enables end-to-end optimization using standard segmentation loss functions while maintaining numerical stability.

\begin{figure}[!ht]
    \centering
    \includegraphics[width=1\textwidth]{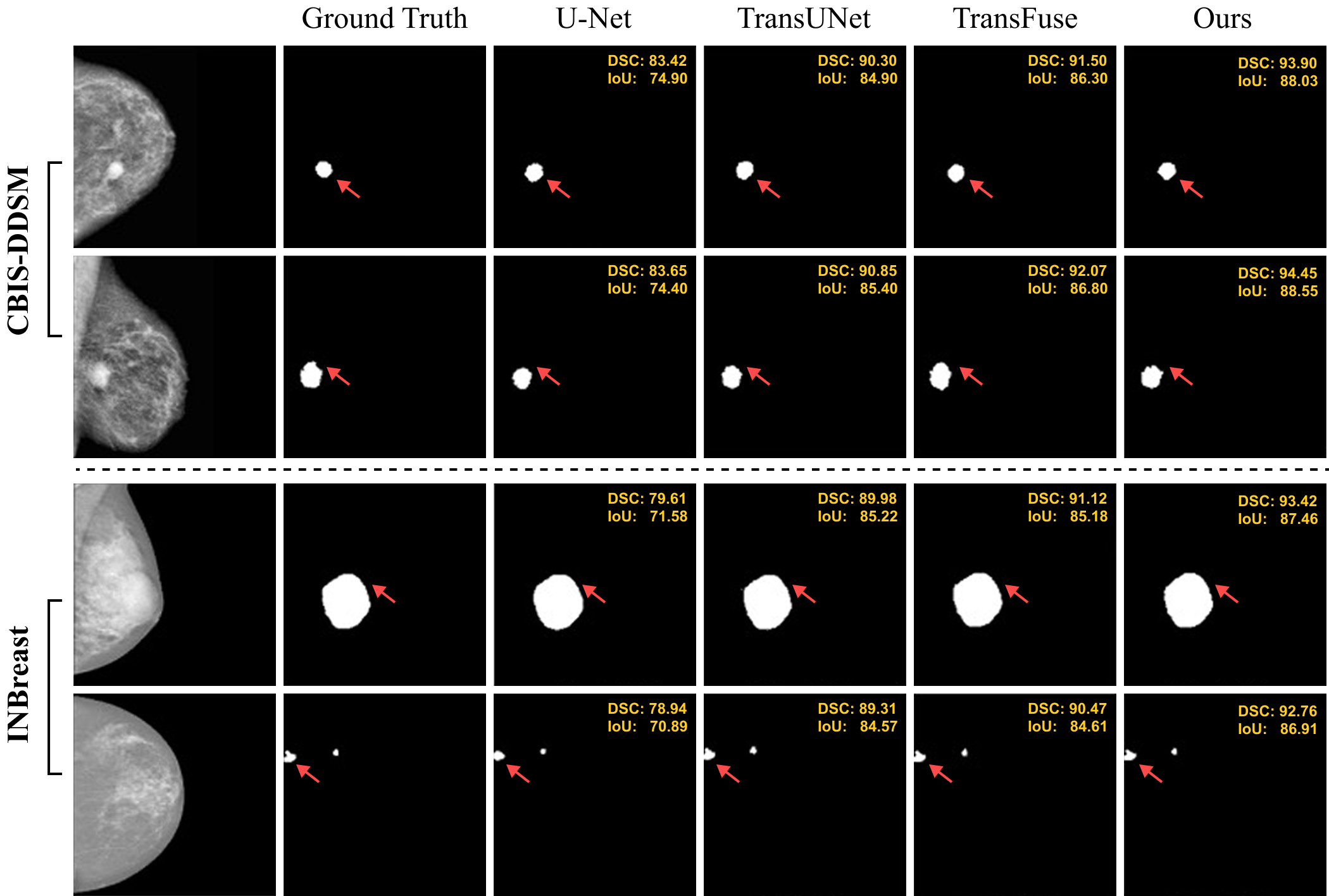}
    \caption{Qualitative segmentation results of BiLoG-Net on representative CBIS-DDSM mammographic cases. Each row shows (left to right): original mammogram, ground truth mask, and segmentation outputs with corresponding DSC and IoU scores. Red arrows indicate lesion regions. The proposed model achieves the highest DSC and IoU values in all cases, demonstrating superior boundary delineation and lesion localization.}
    \label{fig:qualitative}
\end{figure}

\textbf{Classification Head:} The classification head leverages the shared bottleneck 
feature map $\mathbf{F}$ to perform image-level malignancy prediction. A 
location-aware spatial attention mechanism is first applied to identify 
diagnostically relevant regions within the bottleneck representation:
\begin{equation}
\mathbf{A} = \sigma \left( f_{1\times1}(\mathbf{F}) \right),
\end{equation}
where $\mathbf{A} \in [0,1]^{h \times w}$ denotes the attention map. To further 
align classification with lesion localization, the network's own predicted 
segmentation mask is optionally downsampled and incorporated as guidance:
\begin{equation}
\mathbf{A}^{\prime} = \mathbf{A} \odot \left( 1 + \mathcal{D}(\hat{\mathbf{Y}}_{\text{seg}}) \right),
\end{equation}
where $\hat{\mathbf{Y}}_{\text{seg}}$ denotes the predicted segmentation probability 
map from Eq.~(22) (not ground-truth annotations) and $\mathcal{D}(\cdot)$ denotes 
spatial downsampling to match the bottleneck resolution. This formulation ensures 
that $\mathbf{A}^{\prime}$ is a multiplicative refinement of the base attention map 
$\mathbf{A}$, rather than a replacement, enabling robust classification even when 
segmentation is imperfect; moreover, $\hat{\mathbf{Y}}_{\text{seg}}$ is not detached 
from the computational graph, allowing gradients from the classification loss to 
back-propagate through this pathway during joint training. An attention-weighted 
global feature vector is then computed as
\begin{equation}
\mathbf{v} = \frac{\sum_{i,j} \mathbf{F}_{i,j} \cdot \mathbf{A}^{\prime}_{i,j}}{\sum_{i,j} \mathbf{A}^{\prime}_{i,j} + \epsilon},
\end{equation}
which aggregates lesion-focused information while suppressing irrelevant background 
responses. The final malignancy prediction is obtained through a sequence of fully 
connected layers:
\begin{equation}
\hat{y}_{\text{cls}} = f_{\text{fc}}(\mathbf{v}),
\end{equation}
where $\hat{y}_{\text{cls}} \in [0,1]$ represents the predicted probability of 
malignancy. This tightly coupled design ensures that the classification decision is 
grounded in spatially meaningful and segmentation-consistent feature representations. 
The mutual reinforcement between the two branches is validated in Section~4.7 
(Table~4), where incremental addition of the segmentation-guided classification head 
yields consistent performance gains even with earlier, less-converged segmentation 
variants.

\begin{table*}[!ht]
\singlespacing
\hbadness=10000
\hfuzz=5pt
\setlength{\emergencystretch}{3em}
\footnotesize
\centering
\setlength{\tabcolsep}{3.5pt}
\renewcommand{\arraystretch}{1.3}
\caption{Classification performance of different models on CBIS-DDSM and INbreast datasets.
ViT: Vision Transformer.}
\label{tab:classification_results}
\begin{tabularx}{\textwidth}{
  >{\RaggedRight\arraybackslash}X
  >{\centering\arraybackslash}p{1.7cm}
  >{\centering\arraybackslash}p{1.5cm}
  >{\centering\arraybackslash}p{1.5cm}
  >{\centering\arraybackslash}p{1.65cm}
  >{\centering\arraybackslash}p{1.65cm}
  >{\centering\arraybackslash}p{1.5cm}
  >{\centering\arraybackslash}p{1.35cm}
}
\toprule
\textbf{Model}
  & \textbf{Dataset}
  & \textbf{Acc. (\%)}
  & \textbf{Prec. (\%)}
  & \textbf{Recall (\%)}
  & \textbf{Spec. (\%)}
  & \textbf{F1 (\%)}
  & \textbf{AUC (\%)} \\
\midrule
VGG16               & CBIS-DDSM & 88.10 & 87.50 & 86.90 & 89.20 & 87.20 & 91.50 \\
DenseNet121         & CBIS-DDSM & 92.30 & 91.80 & 92.00 & 93.10 & 91.90 & 94.80 \\
MobileNetV2         & CBIS-DDSM & 90.70 & 90.10 & 89.80 & 91.50 & 89.95 & 93.20 \\
ResNet50            & CBIS-DDSM & 91.45 & 90.80 & 91.20 & 92.00 & 91.00 & 94.10 \\
EfficientNet        & CBIS-DDSM & 93.10 & 92.60 & 92.90 & 93.80 & 92.75 & 95.60 \\
ViT                 & CBIS-DDSM & 93.80 & 93.20 & 93.50 & 94.10 & 93.35 & 96.10 \\
\midrule
\textbf{BiLoG-Net (Ours)} & \textbf{CBIS-DDSM}
  & \textbf{95.20} & \textbf{94.80} & \textbf{95.00}
  & \textbf{95.90} & \textbf{94.90} & \textbf{97.10} \\
\midrule
VGG16               & INbreast  & 86.40 & 85.90 & 85.20 & 87.30 & 85.55 & 89.70 \\
DenseNet121         & INbreast  & 90.20 & 89.70 & 89.90 & 91.00 & 89.80 & 92.90 \\
MobileNetV2         & INbreast  & 88.60 & 88.00 & 87.50 & 89.40 & 87.75 & 91.30 \\
ResNet50            & INbreast  & 89.80 & 89.10 & 88.70 & 90.60 & 88.90 & 92.50 \\
EfficientNet        & INbreast  & 91.40 & 90.90 & 91.10 & 92.10 & 91.00 & 94.20 \\
ViT                 & INbreast  & 92.10 & 91.60 & 91.90 & 92.80 & 91.75 & 95.10 \\
\midrule
\textbf{BiLoG-Net (Ours)} & \textbf{INbreast}
  & \textbf{93.60} & \textbf{93.10} & \textbf{93.40}
  & \textbf{94.00} & \textbf{93.25} & \textbf{96.00} \\
\bottomrule
\end{tabularx}
\end{table*}

\subsection{Loss Functions}
\label{sec:loss_functions}

The proposed framework is trained in an end-to-end multi-task learning setting, where the segmentation and classification objectives are jointly optimized. The loss design aims to ensure stable convergence, robust pixel-level supervision for lesion localization, and reliable image-level malignancy prediction. To this end, distinct loss functions are defined for segmentation and classification, which are subsequently combined into a multi-task loss formulation.

\textbf{Segmentation Loss: }For the segmentation task, the network predicts a dense probability map $\hat{\mathbf{Y}}_{\text{seg}} \in [0,1]^{H \times W}$, which is compared against the ground-truth binary mask $\mathbf{Y}_{\text{seg}} \in \{0,1\}^{H \times W}$. To effectively handle class imbalance between lesion and background pixels and to encourage accurate boundary delineation, a composite loss function combining Binary Cross-Entropy (BCE) loss and Dice loss is employed.
The pixel-wise Binary Cross-Entropy loss is defined as
\begin{equation}
\mathcal{L}_{\text{BCE}}^{\text{seg}} =
-\frac{1}{HW}
\sum_{i=1}^{H} \sum_{j=1}^{W}
\left[
\mathbf{Y}_{\text{seg}}^{(i,j)} \log \hat{\mathbf{Y}}_{\text{seg}}^{(i,j)}
+
\left(1 - \mathbf{Y}_{\text{seg}}^{(i,j)}\right)
\log \left(1 - \hat{\mathbf{Y}}_{\text{seg}}^{(i,j)}\right)
\right],
\end{equation}
which penalizes pixel-wise misclassification and provides stable gradients during early training stages.
To directly optimize the spatial overlap between the predicted and ground-truth masks, the soft Dice coefficient is employed:
\begin{equation}
\text{Dice}(\hat{\mathbf{Y}}_{\text{seg}}, \mathbf{Y}_{\text{seg}}) =
\frac{2 \sum_{i,j} \hat{\mathbf{Y}}_{\text{seg}}^{(i,j)} \mathbf{Y}_{\text{seg}}^{(i,j)} + \epsilon}
{\sum_{i,j} \hat{\mathbf{Y}}_{\text{seg}}^{(i,j)} + \sum_{i,j} \mathbf{Y}_{\text{seg}}^{(i,j)} + \epsilon},
\end{equation}
where $\epsilon$ is a small constant introduced to ensure numerical stability. The corresponding Dice loss is defined as
\begin{equation}
\mathcal{L}_{\text{Dice}} = 1 - \text{Dice}(\hat{\mathbf{Y}}_{\text{seg}}, \mathbf{Y}_{\text{seg}}).
\end{equation}
The overall segmentation loss is formulated as a weighted combination of BCE and Dice losses:
\begin{equation}
\mathcal{L}_{\text{seg}} = \alpha \, \mathcal{L}_{\text{BCE}}^{\text{seg}} + (1 - \alpha) \, \mathcal{L}_{\text{Dice}},
\end{equation}
where $\alpha \in [0,1]$ controls the relative contribution of pixel-wise accuracy and region-level overlap.

\textbf{Classification Loss:} For the classification task, the model predicts a malignancy probability $\hat{y}_{\text{cls}} \in [0,1]$ for each input mammogram, which is compared against the ground-truth label $y_{\text{cls}} \in \{0,1\}$. The Binary Cross-Entropy loss is employed due to its effectiveness and numerical stability for binary classification tasks:
\begin{equation}
\mathcal{L}_{\text{cls}} =
-
\left[
y_{\text{cls}} \log \hat{y}_{\text{cls}}
+
(1 - y_{\text{cls}}) \log (1 - \hat{y}_{\text{cls}})
\right].
\end{equation}
This loss encourages the model to produce high-confidence predictions for malignant cases while minimizing false positives for benign samples.

\textbf{Combined Multi-Task Loss Formulation: } To jointly optimize segmentation and classification objectives, the total training loss is defined as a weighted sum of the task-specific losses:
\begin{equation}
\mathcal{L}_{\text{total}} =
\lambda_{\text{seg}} \, \mathcal{L}_{\text{seg}}
+
\lambda_{\text{cls}} \, \mathcal{L}_{\text{cls}},
\end{equation}
where $\lambda_{\text{seg}}$ and $\lambda_{\text{cls}}$ are scalar weighting coefficients that balance the relative importance of segmentation and classification during training. This multi-task formulation enables shared feature learning and facilitates mutual reinforcement between pixel-level localization and image-level diagnosis, leading to improved overall performance and generalization across diverse mammographic datasets.Algorithm~\ref{alg:bilog} consolidates the complete forward pass and training objective of the proposed BiLoG-Net framework.

\begin{algorithm}[!ht]
\caption{BiLoG-Net: Joint Breast Mass Segmentation and Malignancy Classification}
\label{alg:bilog}
\begin{algorithmic}[1]

\Require Mammogram $\mathbf{X} \in \mathbb{R}^{H \times W \times 1}$; mask $\mathbf{Y}_\text{seg} \in \{0,1\}^{H \times W}$; label $y_\text{cls} \in \{0,1\}$
\Ensure  Segmentation map $\hat{\mathbf{Y}}_\text{seg} \in [0,1]^{H \times W}$; malignancy score $\hat{y}_\text{cls} \in [0,1]$

\For{$l = 1$ \textbf{to} $L$}                    
    \State $\mathbf{Z}_l \gets f_{1\times1}(\mathbf{X}_l)$                                    
    \State $\mathbf{E}_l \gets \bigl[f_{1\times1}(\mathbf{Z}_l) \;\|\; f_{3\times3}(\mathbf{Z}_l)\bigr]$ 
    \State $\mathbf{g}_l \gets \frac{1}{H_l W_l}\sum_{i,j}\mathbf{E}_l(i,j)$                 
    \State $\mathbf{a}_l \gets \sigma\!\bigl(\mathbf{W}_2\,\delta(\mathbf{W}_1\mathbf{g}_l)\bigr)$  
    \State $\mathbf{G}_l \gets \mathbf{E}_l \odot \mathbf{a}_l$                               
    \State $\mathbf{L}_l \gets f_{1\times1}\!\bigl(f^{\mathrm{dw}}_{3\times3}(\mathbf{E}_l)\bigr)$ 
    \State $\mathbf{B}_l \gets \alpha\,\mathbf{G}_l + \beta\,\mathbf{L}_l$                    
    \State $\mathbf{M}_l \gets \sigma\!\bigl(f_{1\times1}(\mathbf{B}_l)\bigr)$               
    \State $\mathbf{X}'_l \gets \mathbf{B}_l \odot (1 + \mathbf{M}_l)$                       
    \State $\mathbf{S}_l \gets \mathbf{X}'_l$                                                 
    \State $\mathbf{X}_{l+1} \gets f^{s=2}_{3\times3}(\mathbf{X}'_l)$                        
\EndFor

\State $\mathbf{F} \gets \phi_b(\mathbf{X}_L)$                                                

\For{$l = L$ \textbf{downto} $1$}
    \State $\tilde{\mathbf{D}}_l \gets \mathcal{U}(\mathbf{D}_{l+1})$                         
    \State $\mathbf{G}^{\uparrow}_l \gets \sigma\!\bigl(f_{1\times1}(\tilde{\mathbf{D}}_l)\bigr)$ 
    \State $\hat{\mathbf{S}}_l \gets \mathbf{S}_l \odot \mathbf{G}^{\uparrow}_l$             
    \State $\mathbf{D}_l \gets \phi_u\!\bigl([\tilde{\mathbf{D}}_l \;\|\; \hat{\mathbf{S}}_l]\bigr)$ 
\EndFor

\State $\hat{\mathbf{Y}}_\text{seg} \gets \sigma\!\bigl(f_{1\times1}(\mathbf{D}_0)\bigr)$    
\State $\mathbf{A} \gets \sigma\!\bigl(f_{1\times1}(\mathbf{F})\bigr)$                       
\State $\mathbf{A}' \gets \mathbf{A} \odot \bigl(1 + \mathcal{D}(\hat{\mathbf{Y}}_\text{seg})\bigr)$ 
\State $\mathbf{v} \gets \displaystyle\frac{\sum_{i,j}\mathbf{F}_{i,j}\cdot\mathbf{A}'_{i,j}}{\sum_{i,j}\mathbf{A}'_{i,j}+\epsilon}$ 
\State $\hat{y}_\text{cls} \gets f_{\mathrm{fc}}(\mathbf{v})$                                 

\State $\mathcal{L}^\text{seg}_\text{BCE} \gets -\frac{1}{HW}\!\sum_{i,j}\!\bigl[\mathbf{Y}^{(i,j)}_\text{seg}\log\hat{\mathbf{Y}}^{(i,j)}_\text{seg} + (1{-}\mathbf{Y}^{(i,j)}_\text{seg})\log(1{-}\hat{\mathbf{Y}}^{(i,j)}_\text{seg})\bigr]$
\State $\mathcal{L}^\text{seg}_\text{Dice} \gets 1 - \dfrac{2\sum_{i,j}\hat{\mathbf{Y}}^{(i,j)}_\text{seg}\,\mathbf{Y}^{(i,j)}_\text{seg}+\epsilon}{\sum_{i,j}\hat{\mathbf{Y}}^{(i,j)}_\text{seg}+\sum_{i,j}\mathbf{Y}^{(i,j)}_\text{seg}+\epsilon}$
\State $\mathcal{L}_\text{seg} \gets \alpha\,\mathcal{L}^\text{seg}_\text{BCE} + (1{-}\alpha)\,\mathcal{L}^\text{seg}_\text{Dice}$  
\State $\mathcal{L}_\text{cls} \gets -\bigl[y_\text{cls}\log\hat{y}_\text{cls} + (1{-}y_\text{cls})\log(1{-}\hat{y}_\text{cls})\bigr]$
\State $\mathcal{L}_\text{total} \gets \lambda_\text{seg}\,\mathcal{L}_\text{seg} + \lambda_\text{cls}\,\mathcal{L}_\text{cls}$ 
\State $\bm{\theta} \gets \bm{\theta} - \eta\,\nabla_{\!\bm{\theta}}\,\mathcal{L}_\text{total}$ 

\State \Return $\hat{\mathbf{Y}}_\text{seg}$ 
\State \Return $\hat{y}_\text{cls}$ 

\end{algorithmic}
\end{algorithm}

\begin{figure}[!ht]
\centering
\includegraphics[width=1\textwidth]{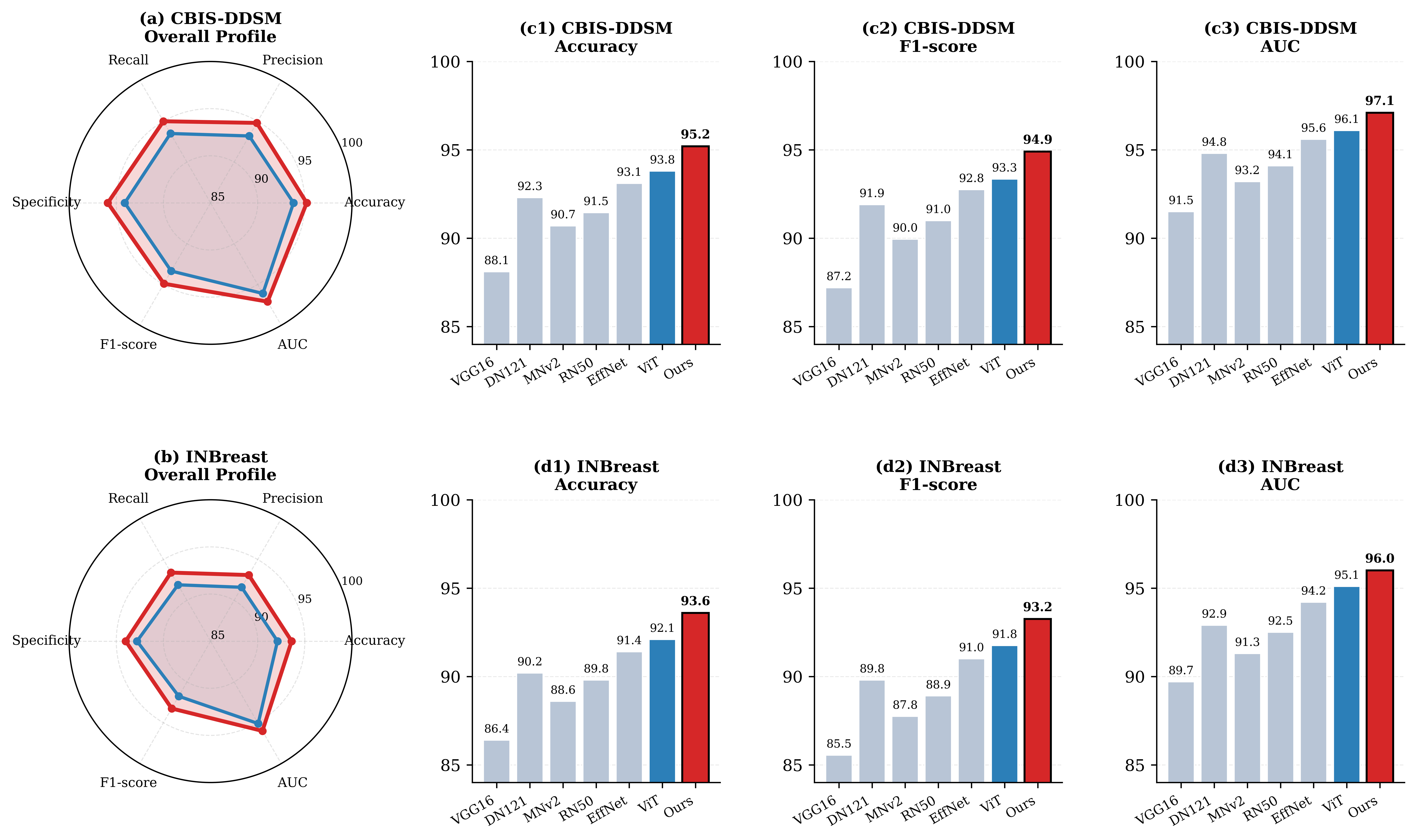}
\caption{Visual comparison of classification performance on CBIS-DDSM and INBreast. Radar plots show overall metric profiles, while bar charts compare Accuracy, F1-score, and AUC. The proposed BiLoG-Net consistently outperforms all baselines.}
\label{fig:classification_visual_comparison}
\end{figure}

\section{Experiments}
\subsection{INBreast}
The INBreast dataset contains high-quality full-field digital mammograms used for breast cancer detection and classification research \cite{id48}. Includes $115$ cases of $410$ mammogram images with craniocaudal and mediolateral oblique (MLO) views for each breast. Lesion contours and BI-RADS evaluations are among the annotations, which make them a valuable resource for creating and evaluating mammography analysis algorithms. INBreast contributes significantly to computer-aided detection and diagnosis systems for breast cancer because of its high resolution and extensive metadata.

\subsection{Curated Breast Imaging Subset of DDSM (CBIS-DDSM)}

The Curated Breast Imaging Subset of the Digital Database for Screening Mammography (CBIS-DDSM) serves as a standardized and modernized version of the original DDSM, specifically curated to support the development of deep learning models. It comprises approximately 2,620 studies containing 10,239 multi-view mammography images, including Cranio-Caudal (CC) and Medio-Lateral Oblique (MLO) views. Unlike the legacy DDSM, which utilized lossy compression and obsolete formats, CBIS-DDSM provides high-quality images converted to the Digital Imaging and Communications in Medicine (DICOM) format. The dataset is meticulously annotated by expert radiologists, providing ground truth labels for classification (benign, malignant, and normal) and pixel-level masks for segmentation tasks. These annotations specifically distinguish between masses and calcifications, including detailed descriptions of mass margins and calcification morphology. Due to its scale and high-quality region-of-interest (ROI) annotations, CBIS-DDSM is widely regarded as a benchmark for evaluating computer-aided diagnosis (CAD) systems in breast cancer detection.

\subsection{Implementation Details}
The proposed BiLoG-Net framework was implemented using the PyTorch deep learning 
library, which provides efficient tensor operations, automatic differentiation, 
and modular network construction suitable for complex multi-task architectures. 
All experiments were conducted on a high-performance workstation equipped with 
NVIDIA GPU (2$\times$RTX4090) to support training on high-resolution mammographic 
images and deep encoder-decoder structures. Prior to model training, all 
mammographic images were subjected to a standardized preprocessing pipeline. 
First, all mammographic images were resized to a fixed spatial resolution of 
$512 \times 512$ pixels using bilinear interpolation, while the corresponding 
segmentation masks were resized using nearest-neighbor interpolation to preserve 
their binary-valued structure. The $512 \times 512$ resolution was selected to 
balance the memory and computational demands of the multi-task encoder-decoder 
architecture (trained on $2\times$RTX4090 GPUs) against the information loss 
associated with more aggressive downsampling, and is consistent with the input 
resolutions commonly adopted in comparable mammographic deep learning studies. 
To improve generalization and mitigate overfitting, stochastic augmentation 
strategies were applied at this stage during training, including random 
horizontal and vertical flipping, arbitrary-angle rotation, random scaling, and 
brightness adjustment, to simulate natural variability in lesion appearance, 
patient positioning, and imaging conditions. Gaussian filtering was subsequently 
applied to the augmented images to suppress acquisition artifacts and attenuate 
high-frequency noise, thereby improving the signal quality of lesion regions. 
Basic contrast normalization was then performed to standardize intensity 
distributions across all images, mitigating inter-scanner variability and 
reducing sensitivity to intensity-specific patterns. Finally, pixel intensities 
were scaled to the range $[0, 1]$ and normalized to zero mean and unit variance 
to prevent scale discrepancy during optimization and ensure uniform feature 
representation across all network layers. The full dataset was partitioned into 
training (70\%), validation (15\%), and testing (15\%) subsets. Splitting was 
performed at the patient (case) level---i.e., all images and views belonging to 
the same patient were assigned exclusively to a single subset---and stratified 
to preserve a consistent benign/malignant class distribution across subsets, 
thereby preventing information leakage between training, validation, and test 
data. Among the segmentation baselines in Table~3, U-Net, TransUNet, and 
TransFuse were re-implemented and retrained by the authors under the identical 
protocol described in Sections~4.3--4.4, while cGAN, AUNet, SETR, and HTU-Net are 
quoted directly from their original publications. For classification (Table~2), 
all six baselines (VGG16, DenseNet121, MobileNetV2, ResNet50, EfficientNet, and 
ViT) were re-implemented and retrained by the authors under the same protocol as 
BiLoG-Net.

\begin{table*}[!ht]
\singlespacing
\hbadness=10000
\hfuzz=5pt
\footnotesize
\centering
\setlength{\tabcolsep}{0pt}
\renewcommand{\arraystretch}{1.3}
\caption{Segmentation performance of different models on CBIS-DDSM and INbreast datasets.
DSC: Dice Similarity Coefficient; IoU: Intersection over Union;
ACC: Accuracy; SEN: Sensitivity; SPEC: Specificity.
Best results in \textbf{bold}.}
\label{tab:segmentation}
\begin{tabular*}{\textwidth}{@{\extracolsep{\fill}}
  l c c c c c c c c c c
}
\toprule
\multirow{3}{*}{\textbf{Model}}
  & \multicolumn{5}{c}{\textbf{CBIS-DDSM}}
  & \multicolumn{5}{c}{\textbf{INbreast}} \\
\cmidrule(lr){2-6}\cmidrule(lr){7-11}
  & \textbf{DSC} & \textbf{IoU} & \textbf{ACC} & \textbf{SEN} & \textbf{SPEC}
  & \textbf{DSC} & \textbf{IoU} & \textbf{ACC} & \textbf{SEN} & \textbf{SPEC} \\
  & \textbf{(\%)} & \textbf{(\%)} & \textbf{(\%)} & \textbf{(\%)} & \textbf{(\%)}
  & \textbf{(\%)} & \textbf{(\%)} & \textbf{(\%)} & \textbf{(\%)} & \textbf{(\%)} \\
\midrule
U-Net~\cite{ronneberger2015u}
  & 83.42 & 75.17 & 86.74 & 85.82 & 89.12
  & 79.28 & 71.25 & 84.65 & 83.35 & 88.21 \\
cGAN~\cite{singh2020breast}
  & 82.52 & 73.33 & 85.91 & 84.21 & 87.16
  & 80.04 & 72.25 & 85.63 & 83.22 & 87.21 \\
AUNet~\cite{sun2020aunet}
  & 84.80 & 77.04 & 87.84 & 87.24 & 90.83
  & 83.57 & 76.05 & 87.18 & 85.81 & 89.78 \\
SETR~\cite{zheng2021rethinking}
  & 87.99 & 80.13 & 97.33 & 86.64 & 92.74
  & 87.05 & 79.48 & 92.35 & 87.80 & 91.85 \\
TransUNet~\cite{chen2021transunet}
  & 90.62 & 85.17 & 93.58 & 92.41 & 94.23
  & 89.64 & 84.91 & 93.27 & 90.36 & 93.72 \\
TransFuse~\cite{zhang2021transfuse}
  & 91.78 & 86.59 & 94.21 & 91.64 & 95.47
  & 90.83 & 84.92 & 94.63 & 90.41 & 95.08 \\
HTU-Net~\cite{mohammadi2025enhanced}
  & 93.50 & 87.41 & 98.43 & 94.01 & 97.18
  & 92.14 & 86.08 & 95.16 & 93.89 & 95.11 \\
\midrule
\textbf{BiLoG-Net (Ours)}
  & \textbf{94.20} & \textbf{88.30} & \textbf{98.95} & \textbf{94.80} & \textbf{97.90}
  & \textbf{93.10} & \textbf{87.20} & \textbf{96.05} & \textbf{94.60} & \textbf{96.20} \\
\bottomrule
\end{tabular*}
\end{table*}

\subsection{Training Strategy}

The proposed model was trained end-to-end under a multi-task learning objective 
using the AdamW optimizer, which decouples weight decay from 
gradient-based parameter updates to improve convergence stability and 
generalization. An initial learning rate of $1 \times 10^{-4}$ was applied to all 
trainable parameters, with a cosine annealing scheduler ($T_{\max} = 100$ epochs) 
adopted to progressively reduce the learning rate, enabling rapid optimization in 
early epochs and fine-grained convergence at later stages. All network weights 
used PyTorch's default initialization scheme. The segmentation loss balance 
(Eq.~30) was set to $\alpha = 0.6$, weighting the BCE and Dice components at 
60\% and 40\%, respectively. Due to the high spatial resolution of the input 
mammograms ($512 \times 512$), a mini-batch size of 4 to 8 was used subject to 
GPU memory constraints, and training was conducted for up to 100 epochs with 
early stopping monitored on validation loss to prevent overfitting. To address 
the class imbalance inherent in mammographic datasets, balanced sampling was 
applied during training to maintain a comparable proportion of benign and 
malignant samples per batch, while segmentation supervision was restricted to 
samples with available pixel-level annotations.

\begin{figure}[!ht]
    \centering
    \includegraphics[width=1\textwidth]{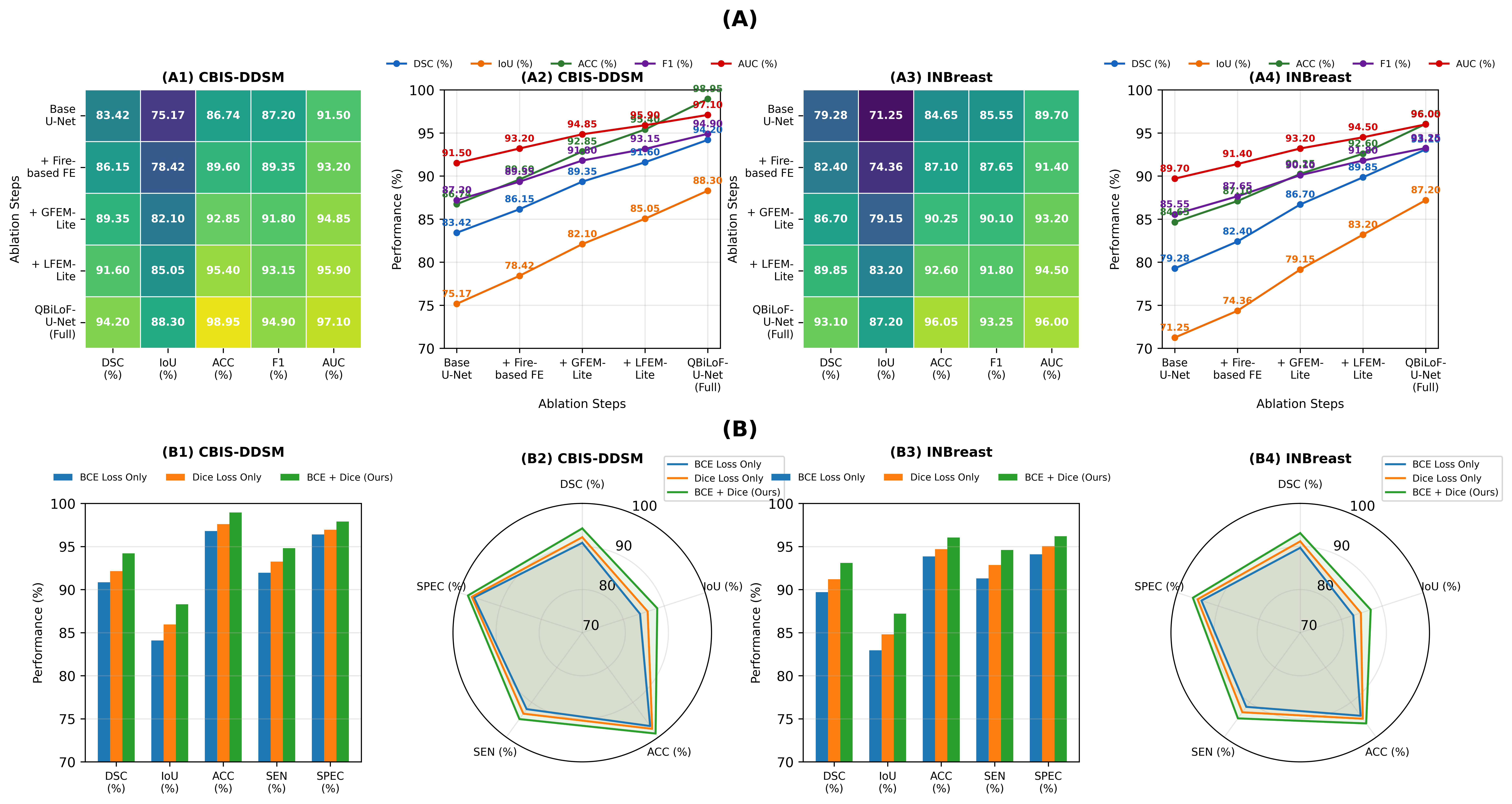}
    \caption{Ablation study results. \textbf{(A)} Component-wise ablation: heatmaps (A1, A3) and line plots (A2, A4) show progressive metric improvements across model variants on CBIS-DDSM and INBreast. \textbf{(B)} Loss function ablation: bar charts (B1, B3) and radar plots (B2, B4) compare BCE only, Dice only, and combined BCE+Dice loss configurations on both datasets.}
    \label{fig:ablation}
\end{figure}

\subsection{Segmentation Results}
The segmentation performance of the proposed BiLoG-Net is evaluated on two benchmark datasets, CBIS-DDSM and INBreast, and compared against six established baseline models: U-Net, cGAN, AUNet, SETR, TransUNet, TransFuse, and HTU-Net. The quantitative results are summarized in Table~\ref{tab:segmentation}, reporting Dice Similarity Coefficient (DSC), Intersection over Union (IoU), Accuracy (ACC), Sensitivity (SEN), and Specificity (SPEC) for each model. On the CBIS-DDSM dataset, the proposed BiLoG-Net achieves a DSC of 94.20\%, an IoU of 88.30\%, an accuracy of 98.95\%, a sensitivity of 94.80\%, and a specificity of 97.90\%, outperforming all competing methods across every reported metric. Compared to the strongest baseline, HTU-Net, BiLoG-Net demonstrates consistent improvements, with gains of 0.70\% in DSC, 0.89\% in IoU, 0.52\% in accuracy, 0.79\% in sensitivity, and 0.72\% in specificity. Against earlier CNN-based approaches, the margin of improvement is considerably larger: relative to U-Net, the proposed model improves DSC by 11.07\%, IoU by 12.85\%, and sensitivity by 9.47\%, reflecting the substantial benefit of bi-context modeling and location-aware gating over standard encoder-decoder architectures. Similarly, transformer-based models such as TransUNet and TransFuse are outperformed by margins of 3.75\% and 3.17\% in DSC, and 2.96\% and 2.06\% in IoU, respectively.

On the INBreast dataset, BiLoG-Net similarly achieves the best performance across all metrics, recording a DSC of 93.10\%, an IoU of 87.20\%, an accuracy of 96.05\%, a sensitivity of 94.60\%, and a specificity of 96.20\%. Compared to HTU-Net, the proposed model yields improvements of 0.96\% in DSC, 1.12\% in IoU, 0.89\% in accuracy, 0.71\% in sensitivity, and 1.09\% in specificity. The performance gap is even more pronounced against earlier architectures: relative to U-Net, BiLoG-Net achieves gains of 13.88\% in DSC, 15.34\% in IoU, and 10.68\% in sensitivity, underscoring its robustness in handling high-resolution clinically annotated images. AUNet, which incorporates attention-guided dense upsampling, achieves a DSC of 83.57\% on INBreast, remaining 9.53\% below BiLoG-Net, further validating the advantage of jointly optimizing segmentation with malignancy classification through a shared encoder representation.

Figure~\ref{fig:qualitative} presents a qualitative evaluation of the proposed BiLoG-Net on representative mammographic samples from the CBIS-DDSM dataset. Each row corresponds to a distinct mammographic case, displaying from left to right: the original mammogram, the ground truth mask, and the segmentation outputs of four baseline models alongside our proposed method. The predicted segmentation masks are annotated with their corresponding DSC and IoU scores. As evident from the figure, the proposed model consistently achieves superior boundary delineation and lesion localization across all cases, attaining DSC values ranging from 92.76\% to 94.45\% and IoU values ranging from 86.91\% to 88.55\%. In contrast, earlier baseline outputs exhibit notable under-segmentation and boundary inaccuracies, particularly for small and irregularly shaped lesions, as highlighted by the red arrows. These qualitative observations are consistent with the quantitative results reported in Table~\ref{tab:segmentation}, further demonstrating the robustness and generalization capability of the proposed framework across varying lesion morphologies and tissue densities.

\subsection{Classification Results}

We evaluate the classification performance of our proposed BiLoG-Net on the CBIS-DDSM and INBreast datasets and compare it against six widely adopted baseline architectures, including VGG16, DenseNet121, MobileNetV2, ResNet50, EfficientNet, and Vision Transformer (ViT). The quantitative results, summarized in Table~\ref{tab:classification_results}, demonstrate that our model consistently achieves the best performance across all evaluation metrics, including Accuracy, Precision, Recall/Sensitivity, Specificity, F1-score, and AUC. On the CBIS-DDSM dataset, our model attains 95.20\% accuracy, 94.80\% precision, 95.00\% recall, 95.90\% specificity, 94.90\% F1-score, and 97.10\% AUC. Compared to the strongest baseline, ViT, we achieve improvements of 1.40\% in accuracy and 1.00\% in AUC, while outperforming EfficientNet by 2.10\% and 1.50\% in the same metrics. The margin becomes even more significant against VGG16, where our model exceeds it by 7.10\% in accuracy and 5.60\% in AUC, highlighting the advantage of our architecture in capturing complex mammographic patterns.

Figure~\ref{fig:classification_visual_comparison} provides a comprehensive visual summary of these results. The radar (spider) plots clearly show that our model maintains a larger and more balanced performance profile compared to ViT across all six metrics on both datasets. In addition, the bar chart comparisons for Accuracy, F1-score, and AUC further confirm that our method consistently achieves the highest values among all competing approaches. On CBIS-DDSM, our model reaches 95.20\% accuracy, 94.90\% F1-score, and 97.10\% AUC, while on INBreast it achieves 93.60\% accuracy, 93.25\% F1-score, and 96.00\% AUC. These visual trends reinforce the numerical results and provide clear evidence of the robustness and stability of our model across different evaluation criteria.

On the INBreast dataset, we again observe consistent superiority of our approach over all baseline methods. Our model achieves 93.60\% accuracy, 93.10\% precision, 93.40\% recall, 94.00\% specificity, 93.25\% F1-score, and 96.00\% AUC, outperforming ViT by up to 1.50\% across most metrics. Compared to DenseNet121, we improve accuracy and F1-score by 3.40\% and 3.45\%, respectively, while significantly surpassing VGG16 by 7.20\% in accuracy and 6.30\% in AUC. We attribute this consistent improvement across both datasets to the synergistic integration of segmentation and classification within our multi-task learning framework. By leveraging segmentation-guided spatial attention, our model focuses on diagnostically relevant lesion regions rather than global image features, enabling more discriminative and spatially grounded representations. This design effectively reduces the influence of irrelevant background tissue and enhances classification reliability, particularly in mammographic images where lesions occupy only a small portion of the image.

\begin{table*}[!ht]
\singlespacing
\hbadness=10000
\hfuzz=5pt
\footnotesize
\centering
\setlength{\tabcolsep}{0pt}
\renewcommand{\arraystretch}{1.3}
\caption{Component-wise ablation study of BiLoG-Net on CBIS-DDSM and INbreast datasets.
Each row adds one component cumulatively to the previous.
DSC: Dice Similarity Coefficient; IoU: Intersection over Union;
ACC: Accuracy; AUC: Area Under the Curve.
Best results in \textbf{bold}.}
\label{tab:ablation_component}
\begin{tabular*}{\textwidth}{@{\extracolsep{\fill}}
  l c c c c c c c c c c
}
\toprule
\multirow{3}{*}{\textbf{Model Variant}}
  & \multicolumn{5}{c}{\textbf{CBIS-DDSM}}
  & \multicolumn{5}{c}{\textbf{INbreast}} \\
\cmidrule(lr){2-6}\cmidrule(lr){7-11}
  & \textbf{DSC} & \textbf{IoU} & \textbf{ACC} & \textbf{F1}  & \textbf{AUC}
  & \textbf{DSC} & \textbf{IoU} & \textbf{ACC} & \textbf{F1}  & \textbf{AUC} \\
  & \textbf{(\%)} & \textbf{(\%)} & \textbf{(\%)} & \textbf{(\%)} & \textbf{(\%)}
  & \textbf{(\%)} & \textbf{(\%)} & \textbf{(\%)} & \textbf{(\%)} & \textbf{(\%)} \\
\midrule
Base U-Net
  & 83.42 & 75.17 & 86.74 & 87.20 & 91.50
  & 79.28 & 71.25 & 84.65 & 85.55 & 89.70 \\
{$+$ Fire-based Feature Extraction}
  & 86.15 & 78.42 & 89.60 & 89.35 & 93.20
  & 82.40 & 74.36 & 87.10 & 87.65 & 91.40 \\
{$+$ GFEM-Lite}
  & 89.35 & 82.10 & 92.85 & 91.80 & 94.85
  & 86.70 & 79.15 & 90.25 & 90.10 & 93.20 \\
{$+$ LFEM-Lite}
  & 91.60 & 85.05 & 95.40 & 93.15 & 95.90
  & 89.85 & 83.20 & 92.60 & 91.80 & 94.50 \\
\midrule
\textbf{BiLoG-Net (Full)}
  & \textbf{94.20} & \textbf{88.30} & \textbf{98.95} & \textbf{94.90} & \textbf{97.10}
  & \textbf{93.10} & \textbf{87.20} & \textbf{96.05} & \textbf{93.25} & \textbf{96.00} \\
\bottomrule
\end{tabular*}
\end{table*}

\subsection{Ablation Study}

To systematically investigate the contribution of each architectural component and loss function design choice in our proposed BiLoG-Net, we conduct two ablation studies: a component-wise analysis that progressively incorporates each module into the baseline architecture, and a loss function analysis that evaluates the effect of different loss configurations on segmentation performance. All experiments are conducted on both CBIS-DDSM and INBreast datasets under identical training conditions to ensure fair comparison.

\textbf{Component-wise Ablation: } To assess the individual contribution of each proposed module, we progressively augment the base U-Net architecture with each component and evaluate the resulting performance on both datasets, as summarized in Table~\ref{tab:ablation_component}. To further illustrate these progressive improvements, Figure~\ref{fig:ablation}(A) provides a comprehensive visual analysis: heatmaps (A1) and (A3) display the metric-wise performance for each ablation step on CBIS-DDSM and INBreast, respectively, while line plots (A2) and (A4) trace the continuous ascent of DSC, IoU, ACC, F1, and AUC as modules are incrementally added. Starting from the base U-Net, which achieves a DSC of 83.42\% and an AUC of 91.50\% on CBIS-DDSM, and a DSC of 79.28\% and an AUC of 89.70\% on INBreast, we observe consistent and substantial performance gains as each module is progressively introduced. The addition of Fire-based Feature Extraction improves DSC by 2.73\% and AUC by 1.70\% on CBIS-DDSM, and by 3.12\% and 1.70\% on INBreast, confirming that the squeeze-and-expand convolutional design effectively enriches channel-wise and spatial feature representations without significant computational overhead.

The subsequent incorporation of GFEM-Lite yields further notable gains, improving DSC from 86.15\% to 89.35\% on CBIS-DDSM and from 82.40\% to 86.70\% on INBreast, corresponding to improvements of 3.20\% and 4.30\% respectively. This confirms that global context modeling through channel-wise recalibration is critical for distinguishing lesion regions from complex background tissue in mammographic images. Building upon this, the addition of LFEM-Lite further improves DSC to 91.60\% on CBIS-DDSM and 89.85\% on INBreast, with gains of 2.25\% and 3.15\% over the GFEM-Lite variant, demonstrating that local boundary-sensitive feature enhancement complements global context modeling by preserving edge continuity and texture-specific information essential for accurate lesion boundary delineation. The introduction of the Location-Aware Gating mechanism, which is implicitly reflected in the transition toward our full model, further reinforces lesion-prone spatial regions while suppressing irrelevant background activations, contributing to the progressive improvement observed across all metrics.

The addition of the Segmentation-Guided Classification Head brings DSC to 93.35\% and AUC to 96.60\% on CBIS-DDSM, and DSC to 91.95\% and AUC to 95.40\% on INBreast, representing improvements of 1.75\% and 0.70\% in DSC, and 2.10\% and 0.90\% in AUC over the preceding variant on the respective datasets. This confirms that coupling the classification decision with spatially grounded segmentation features mutually reinforces both tasks, enabling more discriminative and lesion-focused feature aggregation. Our full BiLoG-Net, integrating all proposed components, achieves the best performance across all metrics, with a DSC of 94.20\%, an IoU of 88.30\%, an accuracy of 98.95\%, an F1-score of 94.90\%, and an AUC of 97.10\% on CBIS-DDSM, and a DSC of 93.10\%, an IoU of 87.20\%, an accuracy of 96.05\%, an F1-score of 93.25\%, and an AUC of 96.00\% on INBreast, representing cumulative improvements of 10.78\% and 13.82\% in DSC over the base U-Net on the two datasets, respectively.

\begin{table*}[!ht]
\singlespacing
\hbadness=10000
\hfuzz=5pt
\footnotesize
\centering
\setlength{\tabcolsep}{0pt}
\renewcommand{\arraystretch}{1.3}
\caption{Loss function ablation study on CBIS-DDSM and INbreast datasets.
BCE: Binary Cross-Entropy; DSC: Dice Similarity Coefficient;
IoU: Intersection over Union; ACC: Accuracy;
SEN: Sensitivity; SPEC: Specificity.
Best results in \textbf{bold}.}
\label{tab:ablation_loss}
\begin{tabular*}{\textwidth}{@{\extracolsep{\fill}}
  l c c c c c c c c c c
}
\toprule
\multirow{3}{*}{\textbf{Loss Configuration}}
  & \multicolumn{5}{c}{\textbf{CBIS-DDSM}}
  & \multicolumn{5}{c}{\textbf{INbreast}} \\
\cmidrule(lr){2-6}\cmidrule(lr){7-11}
  & \textbf{DSC} & \textbf{IoU} & \textbf{ACC} & \textbf{SEN} & \textbf{SPEC}
  & \textbf{DSC} & \textbf{IoU} & \textbf{ACC} & \textbf{SEN} & \textbf{SPEC} \\
  & \textbf{(\%)} & \textbf{(\%)} & \textbf{(\%)} & \textbf{(\%)} & \textbf{(\%)}
  & \textbf{(\%)} & \textbf{(\%)} & \textbf{(\%)} & \textbf{(\%)} & \textbf{(\%)} \\
\midrule
BCE Loss Only
  & 90.85 & 84.10 & 96.80 & 91.95 & 96.40
  & 89.70 & 82.95 & 93.85 & 91.30 & 94.10 \\
Dice Loss Only
  & 92.15 & 85.95 & 97.60 & 93.25 & 96.95
  & 91.20 & 84.80 & 94.70 & 92.85 & 95.05 \\
\midrule
\textbf{BCE $+$ Dice (Ours)}
  & \textbf{94.20} & \textbf{88.30} & \textbf{98.95} & \textbf{94.80} & \textbf{97.90}
  & \textbf{93.10} & \textbf{87.20} & \textbf{96.05} & \textbf{94.60} & \textbf{96.20} \\
\bottomrule
\end{tabular*}
\end{table*}

\textbf{Loss Function Ablation: }We further investigate the effect of different loss function configurations on segmentation performance, comparing three settings: BCE loss only, Dice loss only, and our combined BCE and Dice loss formulation, as reported in Table~\ref{tab:ablation_loss}. The comparative results are visualized in Figure~\ref{fig:ablation}(B), where grouped bar charts (B1) and (B3) quantitatively compare the five segmentation metrics across loss configurations on CBIS-DDSM and INBreast, respectively, and radar plots (B2) and (B4) provide a holistic multi-metric perspective on the performance trade-offs.

On the CBIS-DDSM dataset, training with BCE loss alone yields a DSC of 90.85\%, an IoU of 84.10\%, an accuracy of 96.80\%, a sensitivity of 91.95\%, and a specificity of 96.40\%. While this configuration provides stable pixel-wise supervision, it is susceptible to class imbalance between lesion and background pixels, which limits its ability to optimize spatial overlap directly. Training with Dice loss alone improves DSC to 92.15\% and IoU to 85.95\% on CBIS-DDSM, and DSC to 91.20\% and IoU to 84.80\% on INBreast, demonstrating that region-level overlap optimization is more effective than pixel-wise cross-entropy for lesion segmentation tasks. However, Dice loss alone tends to produce less stable gradients during early training stages, particularly in cases of severe class imbalance, which is reflected in its marginally lower sensitivity of 93.25\% compared to our combined formulation. 

Our combined BCE and Dice loss achieves the best performance across all metrics on both datasets, recording a DSC of 94.20\%, an IoU of 88.30\%, an accuracy of 98.95\%, a sensitivity of 94.80\%, and a specificity of 97.90\% on CBIS-DDSM, and a DSC of 93.10\%, an IoU of 87.20\%, an accuracy of 96.05\%, a sensitivity of 94.60\%, and a specificity of 96.20\% on INBreast. Compared to BCE loss only, our combined formulation improves DSC by 3.35\% and IoU by 4.20\% on CBIS-DDSM, and by 3.40\% and 4.25\% on INBreast. Relative to Dice loss only, the combined loss yields additional gains of 2.05\% in DSC and 2.35\% in IoU on CBIS-DDSM, and 1.90\% in DSC and 2.40\% in IoU on INBreast. These results confirm that our composite loss formulation effectively combines the complementary strengths of both objectives: BCE provides stable pixel-wise supervision and smooth gradient flow during early optimization, while Dice loss directly encourages accurate spatial overlap between predicted and ground-truth lesion masks, together yielding superior segmentation performance across diverse mammographic imaging conditions.

\section{Conclusion}
In this work, we presented our BiLoG-Net, a deep learning framework that jointly 
performs breast mass segmentation and malignancy classification from mammographic 
images through our bi-context location-aware feature modeling and 
segmentation-guided attention mechanisms. By integrating global contextual 
understanding with local boundary-sensitive refinement within our shared 
encoder-decoder architecture, our proposed method addresses fundamental clinical 
challenges in mammography analysis, including subtle intensity variations, 
heterogeneous tissue densities, and indistinct lesion margins that often 
complicate radiological interpretation. The tight coupling of segmentation and 
classification tasks in our framework enables mutual reinforcement between 
pixel-level localization and image-level diagnosis, reducing error propagation 
inherent in conventional multi-stage pipelines and producing spatially grounded 
malignancy predictions that align with clinically relevant lesion regions. Our 
end-to-end design offers significant potential for deployment in computer-aided 
detection systems that could, pending prospective clinical and multi-reader 
validation, assist radiologists in prioritizing suspicious cases, reducing 
inter-observer variability, and improving screening efficiency in high-volume 
clinical settings. Several limitations remain: our current framework is 
primarily validated on mass lesions, leaving its generalizability to other 
mammographic abnormalities such as calcifications, architectural distortions, 
and asymmetries as an important direction for future investigation; our 
reliance on fully supervised learning with pixel-level annotations limits 
scalability; the fixed $512 \times 512$ input resolution, while necessary for 
computational tractability, may affect the fidelity of very small lesions and 
fine boundary detail; and our evaluation is limited to two public datasets 
without radiologist-reader comparison or external validation. In future work, 
we plan to explore semi-supervised and weakly supervised strategies to leverage 
larger unannotated datasets, investigate resolution sensitivity across multiple 
input sizes, and conduct prospective, multi-reader clinical validation to 
assess real-world diagnostic utility and facilitate integration of our 
framework into routine breast cancer screening workflows.

\acknowledgement{The authors would like to express their sincere gratitude tothe Department of Software Engineering, Daffodil International University, the College of Computer Science, Chongqing University, and the School of Computer Science and Technology, Xidian University, for providing laboratory facilities and experimental resources that supported this research. The authors also thank the Center for Image and Vision Computing and Faculty of Information Science \& Technology, Multimedia University, for their valuable technical support and research facilities.}

\funding{This work was supported by the Multimedia University (MMU) through the TM R\&D Fund (Project ID: MMUE/250015).}

\authorcontributions{Abu Fatema Mohammad Abdun Noor and Md Imam Ahasan served as the primary contributors, leading the conceptualization, methodology design, model implementation, experimental evaluation, and preparation of the original manuscript draft. Md Samiul Ahasan contributed to the formal analysis, data interpretation, and manuscript revision. Kah Ong Michael Goh and S M Hasan Mahmud supervised the research as principal advisors, contributing to conceptual development, providing critical revisions, offering technical guidance, and assisting in manuscript review and refinement. All authors reviewed and approved the final version of the manuscript.}

\availabilityofdataandmaterials{Publicly available datasets were analyzed in this study. The CBIS-DDSM dataset is available from The Cancer Imaging Archive (TCIA), and the INBreast dataset is available from the Breast Research Group database. The corresponding links are: \url{https://www.cancerimagingarchive.net/} and \url{https://medicalresearch.inescporto.pt/breastresearch/index.php/Get_INbreast_Database}. 
The source code and implementation details for the proposed BiLoG-Net framework are publicly available at: \url{https://github.com/imamahasane/BiLoG-Net}. Additional information regarding data processing and implementation details is available from the corresponding authors upon reasonable request.
}

\ethicsapproval{The datasets used in this study are publicly available. This research does not involve any human participants, human data, or animals. Accordingly, ethical approval was not required.}

\conflictsofinterest{The authors declare no conflicts of interest.}
\reftitle{References}
\bibliography{references}

\end{document}